%% file: aaai23.tex
\documentclass[letterpaper]{article} 
\usepackage[submission]{aaai23}  
\usepackage{times}  
\usepackage{helvet}  
\usepackage{courier}  
\usepackage[hyphens]{url}  
\usepackage{graphicx} 
\usepackage{natbib}  
\usepackage{caption} 
\usepackage[table,xcdraw]{xcolor}
\usepackage{tikz}
\usepackage{comment}
\usepackage{amsmath,amssymb} 
\usepackage{color}
\usepackage{multirow}
\usepackage{algorithm}
\usepackage{algorithmic}
\usepackage{adjustbox}

\usepackage{newfloat}
\usepackage{listings}
\DeclareCaptionStyle{ruled}{labelfont=normalfont,labelsep=colon,strut=off} 
\floatstyle{ruled}
\newfloat{listing}{tb}{lst}{}
\floatname{listing}{Listing}
\usepackage[linkcolor=pink]{hyperref}

\author{
    Minh-Long Luu\textsuperscript{\rm 1} \quad
    Zeyi Huang\textsuperscript{\rm 2} \quad
    Eric P. Xing\textsuperscript{\rm 3} \quad
    Yong Jae Lee\textsuperscript{\rm 2} \quad
    Haohan Wang\textsuperscript{\rm 4} \quad
}
\affiliations{
    \textsuperscript{\rm 1} International University-VNUHCM\quad
    \textsuperscript{\rm 2} University of Wisconsin-Madison\\
    \textsuperscript{\rm 3} Carnegie Mellon University\quad
    \textsuperscript{\rm 4} University of Illinois Urbana-Champaign\\
    minhlong9413@gmail.com,
    zhuang479@wisc.edu,
    epxing@cs.cmu.edu,
    yongjaelee@cs.wisc.edu,
    haohanw@illinois.edu
    
}
\title{Expeditious Saliency-guided Mix-up through Random Gradient Thresholding}

\usepackage{bibentry}

\newcommand{\mixup}{Input Mix-up}
\newcommand{\cutmix}{{CutMix}}
\newcommand{\puzzlemix}{{PuzzleMix}}
\newcommand{\comixup}{{Co-Mixup}}
\newcommand{\manifoldmixup}{{Manifold Mixup}}
\newcommand{\saliencymix}{{SaliencyMix}}
\newcommand{\rrlmix}{{R-Mix}}
\newcommand{\cifar}{CIFAR-100}
\newcommand{\imagenet}{ImageNet}

\begin{document}

\maketitle

\begin{abstract}
\input{secs/abstract.tex}
\end{abstract}

\section{Introduction}
\label{sec:intro}
\input{secs/intro}

\section{Background and Motivation}
\label{sec:background}
In this Section, we provide background knowledge about mix-up training, and empirical results serving as motivation for our method.
\input{secs/background}

\section{R-Mix: An Expeditious Saliency-based Mixup}
\label{sec:method}
\input{secs/method.tex}

\section{Experiments}
\label{sec:exp}
In this Section, we describe the datasets, models and training pipelines to benchmark our method on for different tasks: Image Classification, Weakly Supervised Object Localization, Expected Calibration Error, and Robustness to Adversarial Attack.
\input{secs/experiment.tex}

\section{Related Work}
\label{sec:related}
\input{secs/related.tex}

\section{Conclusion}
\label{sec:concl}
In this paper, we show that randomization is capable of performing at the cutting-edge tier, suggesting an unexplored domain in recent advances of mix-up research. Driven by the effectiveness of a mix-up research path over one another, we propose \rrlmix{}, a simple training heuristic that lies at the junction of the two routes. Extensive experiments on image classification, weakly supervised object localization, calibration, and robustness to the adversarial attack show a consistent improvement or on-par performance with state-of-the-art methods while offering speed and simplicity of Arbitrary Mix-up.
Finally, we describe RL-Mix, an early experiment of a Reinforcement Learning - powered agent to automatically decides the mixing regions based on the performance of the classifier, which has shown a competitive capability on \cifar{}, laying the foundation of low-effort hyperparameter tuning mix-up.

\section*{Acknowledgement}
This work was supported in part by the Institute of Information \& communications Technology Planning \& Evaluation (IITP) grant funded by the Korea government (MSIT) (No. 2022- 0-00871, Development of AI Autonomy and Knowledge Enhancement for AI Agent Collaboration).

\bibliography{aaai23}

\clearpage

\section*{Mix-up Methods Summary}
In this Section, we provide brief summary of some mix-up methods. We refer readers to the survey paper \cite{naveed2021survey} for a more detailed overview.

\begin{itemize}
    \item Input Mix-up \cite{zhang2018mixup}: first variant, simply interpolates two samples based on the weight sampled from the Beta distribution, then train the Deep Neural Network with the mixed samples.
    \item Manifold Mix-up \cite{verma2019manifold} interpolates the two samples at a random layer in the DNN instead of the first layer.
    \item CutMix \cite{yun2019cutmix} selects a random rectangular region with size sampled from the Beta distribution from one image and pastes it and paste it to another.
    \item SaliencyMix \cite{uddin2021saliencymix} works similar to CutMix, but it selects the top salient regions based on the saliency calculation.
    \item PuzzleMix \cite{kim2020puzzlemix} first calculates the saliency of the image, then optimizes secondary objectives to ensure rich supervisory signal of the mixed image.
    \item Co-Mixup \cite{kim2021comixup} extends PuzzleMix to the batch level instead of a pair of images by optimizing mixing objectives for the whole batch. 
    \item AutoMix \cite{mixup10} separates mixing and classifying into separate parts and adds encoders to the training pipeline to automatically mix images and labels.
    \item AlignMix \cite{mixup9} calculates the distance of feature vectors, measures assignment matrix using Sinkhorn-Knopp algorithm, then mixes the images based on the matrix.
\end{itemize}

\begin{table}[h!]
\centering
\begin{tabular}{lcccc}
\hline
 & Model & Eps & Cost/Eps & Accuracy \\ \hline
 Vanilla & PARN-18 & 300 & 1/1 & 76.41 \\
Input & PARN-18 & 300 & 1/1 & 77.57 \\
Manifold & PARN-18 & 300 & 1/1 & 78.36 \\
CutMix+ & PARN-18 & 300 & 1/1 & 80.60 \\
SaliencyMix & PARN-18 & 2000 & 2/1 & 80.31 \\
PuzzleMix & PARN-18 & 300 & 2.9/1 & 79.38 \\
Co-Mixup & PARN-18 & 300 & 3/1 & 80.13 \\
AutoMix & RN-18 & 1200 & 2/1 & 80.95 \\
AlignMix & PARN-18 & 2000 & 1.05/1 & 81.71 \\ \hline
R-Mix & PARN-18 & 300 & 2/1 & 81.49 \\ \hline
\end{tabular}
\caption{Model, number of Epoch, Cost per Epoch, and Accuracy of various mix-up methods on CIFAR-100. PARN: PreActResNet. RN: ResNet.}
\label{tab:cifar}
\end{table}

We provide comparisons of \rrlmix{} with other mix-up methods in Table \ref{tab:cifar} and \ref{tab:imagenet}. 

\begin{table}[h!]
\centering
\begin{tabular}{lcccc}
\hline 
 & Model & Eps & Cost/Eps & Accuracy \\ \hline 
Vanilla & RN-50 & 100 & 1/1 & 75.97 \\
Input & RN-50 & 100 & 1/1 & 77.03 \\
Manifold & RN-50 & 100 & 1/1 & 76.70 \\
CutMix & RN-50 & 100 & 1/1 & 77.08 \\
SaliencyMix & RN-50 & 100 & 2/1 & 77.14 \\
PuzzleMix & RN-50 & 100 & 2.9/1 & 77.51 \\
Co-Mixup & RN-50 & 100 & 3/1 & 77.61 \\
AutoMix & RN-50 & 100 & 2/1 & 77.91 \\
AlignMix & RN-50 & 100 & 1.05/1 & 78.00 \\ \hline 
R-Mix & RN-50 & 100 & 2/1 & 77.41 \\ \hline 
\end{tabular}
\caption{Model, number of Epoch, Cost per Epoch and Accuracy of various mix-up methods on ImageNet. PARN: PreActResNet. RN: ResNet. Note that at the time of writing, AlignMix has not released the 100 epochs training code nor the models for it.}
\label{tab:imagenet}
\end{table}

\section*{Ablation Study: different ways to randomly mix images}
In this section, we describe the design process that leads to the current implementation of R-Mix. We summarize the design steps in Table \ref{tab:study}. Experiments are conducted on CIFAR-100 \cite{Krizhevsky09cifar100} using PreActResNet-18 \cite{he2016preact}.

First, we apply the cut-and-paste strategy to the top salient regions of the image (analogy to SaliencyMix \cite{uddin2021saliencymix}, Strategy 1). The top salient region is randomly selected from the top-k value described in the main paper. We observe that it offers marginal improvement (0.5\%) over PuzzleMix \cite{kim2020puzzlemix}.

Second, we apply cut-and-paste strategy to both the top and least salient regions, and keep the rest intact (Strategy 2). This strategy reduces the accuracy of the model by 0.36\%.

Third, we apply mixing to the top-top and least-least salient regions, and select only the top patches in the top-least case (Strategy 3). This is the proposed R-Mix method. This strategy gives 81.49\% accuracy, which is the best so far.

Finally, we apply the same strategy for the top-top and least-least regions, but select only the least region in the top-least case (Strategy 4). This time, the performance is significantly decreased by up to 6\%.

Seeing that none of the strategy works as good as Strategy 3, we name it R-Mix.

\begin{table*}[h!]
\centering
\begin{tabular}{llccc}
\hline
 &  & Usage & Mix strategy & Accuracy \\ \hline
\multirow{2}{*}{Strategy 1 (SaliencyMix)} & Top Salient & Yes & Cut-and-Paste & \multirow{2}{*}{79.88\%} \\
 & Least Salient & No & None &  \\ \hline
\multirow{2}{*}{Strategy 2} & Top Salient & Yes & Cut-and-Paste & \multirow{2}{*}{79.52\%} \\
 & Least Salient & Yes & Cut-and-Paste &  \\ \hline
\multirow{3}{*}{Strategy 3 (R-Mix)} & Top-Top Salient & Yes & Mix & \multirow{3}{*}{\textbf{81.49\%}} \\
 & Least-Least Salient & Yes & Mix &  \\
 & Top-Least Salient & Yes & Select Top only &  \\ \hline
\multirow{3}{*}{Strategy 4} & Top-Top Salient & Yes & Mix & \multirow{3}{*}{75.42\%} \\
 & Least-Least Salient & Yes & Mix &  \\
 & Top-Least Salient & Yes & Select Least only & \\ \hline
\end{tabular}
\caption{Design steps that lead to the implementation of R-Mix. We try different ways to mix images, and then observe Strategy 3 offers the best result. We study on CIFAR-100 using PreActResNet-18.}
\label{tab:study}
\end{table*}

\section*{Implementation details}
\begin{figure*}[t]
  \centering
  \includegraphics[width=2\columnwidth]{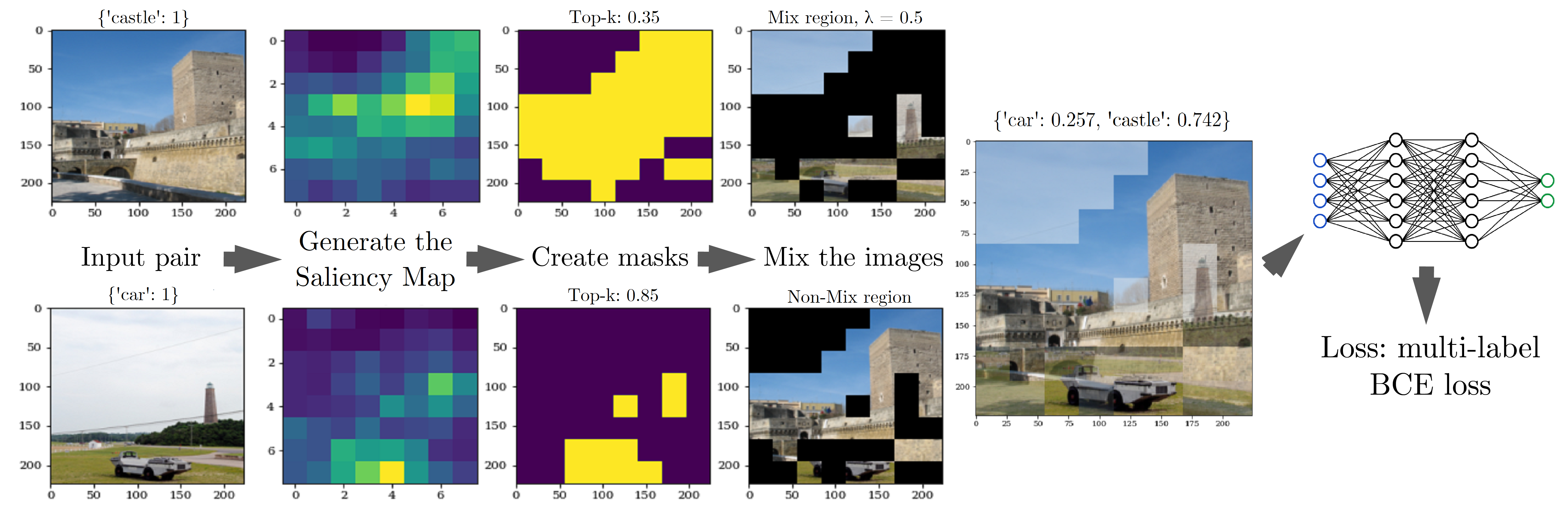}
  \caption{Training pipeline of R-Mix. First, it calculates the saliency map, then divies the map into two regions. Next, it mixes the images based on the region the patch belongs to. Finally, it combines the number of patches and mixing ratio to determine the weights of the inputs.}
  \label{fig:pipeline}
\end{figure*}

We visualize the training pipeline of R-Mix in Figure \ref{fig:pipeline}.  We will release our source code under MIT License upon acceptance. Here, we describe implementation details to reproduce the result:
\begin{itemize}
    \item \textbf{CIFAR-100}. We use four model architectures:
    PreActResNet-18 \cite{he2016preact}, Wide ResNet 16-8 and 28-10 \cite{zagoruyko2017widern}, and ResNeXt 29-4-24 \cite{xie2016resnext}. Wide ResNet models do not use Dropout. All models are trained for 300 epochs, except WRN28-10 is trained for 400 epochs following the original implementation in PuzzleMix \cite{kim2020puzzlemix}. Augmentation includes Random Horizontal Clip and Random Crop with padding 2. Images are normalized channel-wise following well-known mean and standard deviation values.
    We train the models with SGD algorithm using batch size 100, Nesterov Momentum 0.9, and weight decay 0.0001. The OneCycleLR parameters are set as follows: div factor 100, final div factor 10000, max LR 0.3.
    \item \textbf{ImageNet}. For ImageNet \cite{Russakovski2015ImageNet}, we follow the 100 epoch training protocol used in \comixup{} \cite{kim2021comixup}. We keep the training pipeline the same, replacing \comixup{} part with R-Mix.
\end{itemize}

\section*{Sample visualizations}
\begin{figure*}[t]
  \centering
  \includegraphics[width=1\columnwidth]{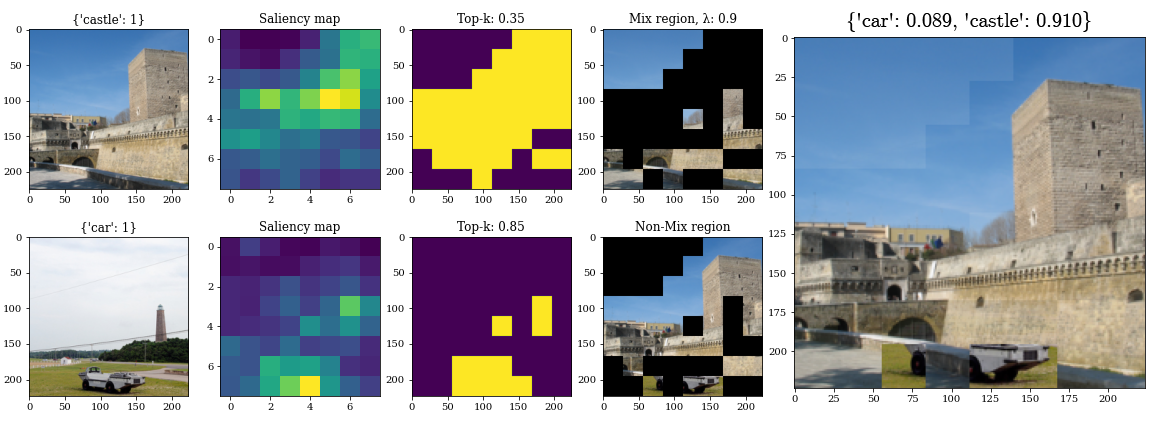}
  \includegraphics[width=1\columnwidth]{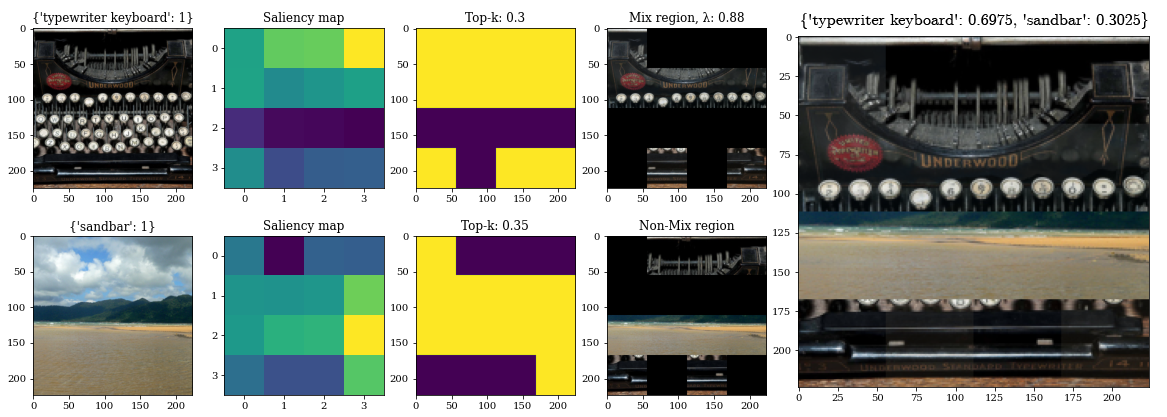}
  \includegraphics[width=1\columnwidth]{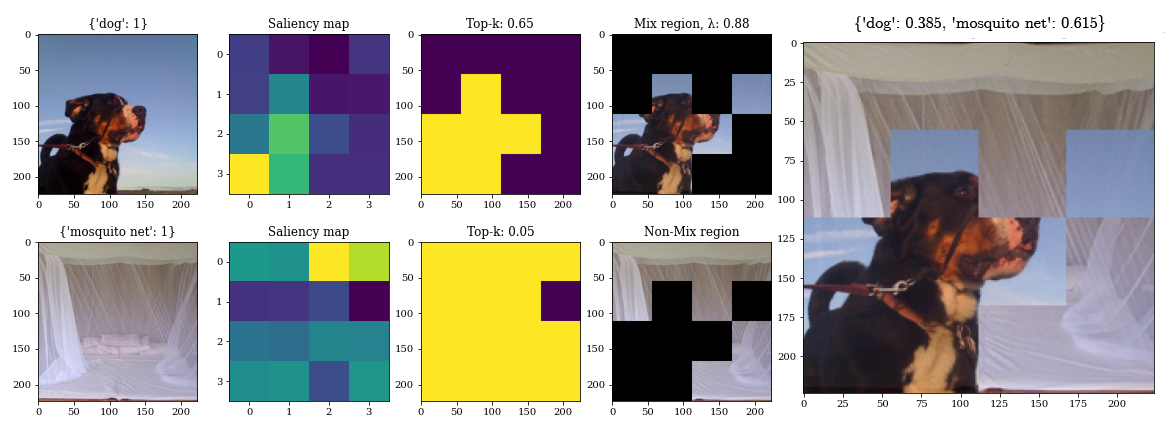}
  \includegraphics[width=1\columnwidth]{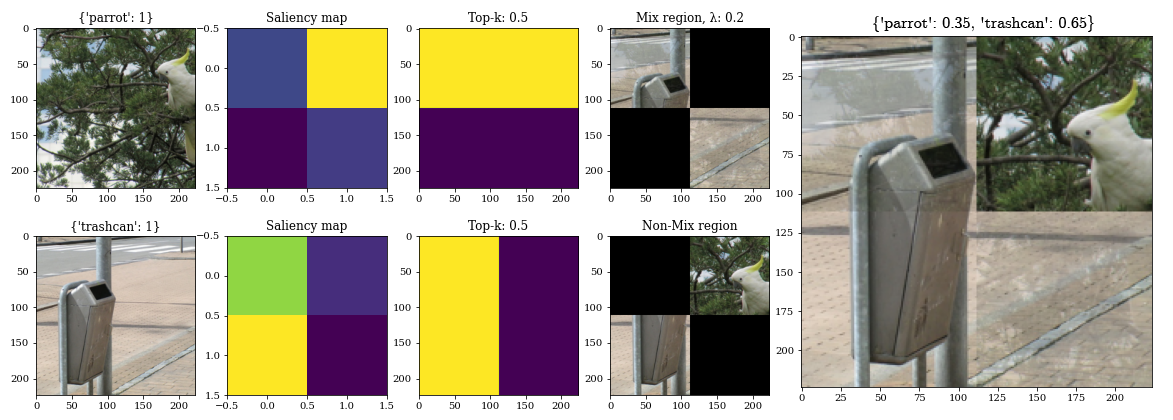}
  \caption{Sample visualization of images produced by R-Mix. }
  \label{fig:mixup234}
\end{figure*}

Sample visualizations are in Figure \ref{fig:mixup234}
\end{document}

%% file: secs/abstract.tex
Mix-up training approaches have proven to be effective in improving the generalization ability of Deep Neural Networks. Over the years, the research community expands mix-up methods into two directions, with extensive efforts to improve saliency-guided procedures but minimal focus on the arbitrary path, leaving the randomization domain unexplored. In this paper, inspired by the superior qualities of each direction over one another, we introduce a novel method that lies at the junction of the two routes. By combining the best elements of randomness and saliency utilization, our method balances speed, simplicity, and accuracy. We name our method \rrlmix{} following the concept of "Random Mix-up". We demonstrate its effectiveness in generalization, weakly supervised object localization, calibration, and robustness to adversarial attacks. Finally, in order to address the question of whether there exists a better decision protocol, we train a Reinforcement Learning agent that decides the mix-up policies based on the classifier's performance, reducing dependency on human-designed objectives and hyperparameter tuning. Extensive experiments further show that the agent is capable of performing at the cutting-edge level, laying the foundation of fully automatic mix-up. Our code is released at \href{https://github.com/minhlong94/Random-Mixup}{\texttt{https://github.com/minhlong94/Random-Mixup}}.

%% file: secs/intro.tex
Mix-up, a data augmentation strategy to increase a deep neural network (DNN)'s predictive performance, 
has drawn a lot of attention in recent years, 
along with the numerous initiatives made to 
pushing various deep learning models to move up the state-of-the-art leaderboard
on multiple benchmarks and different applications. 
The pioneering idea, 
\mixup{}, introduced by \cite{zhang2018mixup},
simply interpolates two samples in a linear manner
and has been proven to play a significant role 
in improving a model's predictive performance
with hardly any additional computing cost. Recently, theoretical explanations for how \mixup{} enhances robustness and generalization have been studied \cite{zhang2021howmixup}.

Building upon the empirical success of these mix-up methods, 
the community has explored multiple directions to further 
improve the mix-up idea. \manifoldmixup{} \cite{verma2019manifold} extends the original mix-up by mixing at a random layer in the model. {AugMix} \cite{hendrycks*2020augmix} first augments the images by different combinations of augmentation techniques, then finally mixes them together to increase the robustness of DNNs. 
\cutmix{} \cite{yun2019cutmix} uses a spatial copy-and-paste-based strategy on other samples to create the new mixed-up sample
and has also been used widely in various applications.

\begin{figure}[t]
  \centering
  \includegraphics[width=0.95\columnwidth]{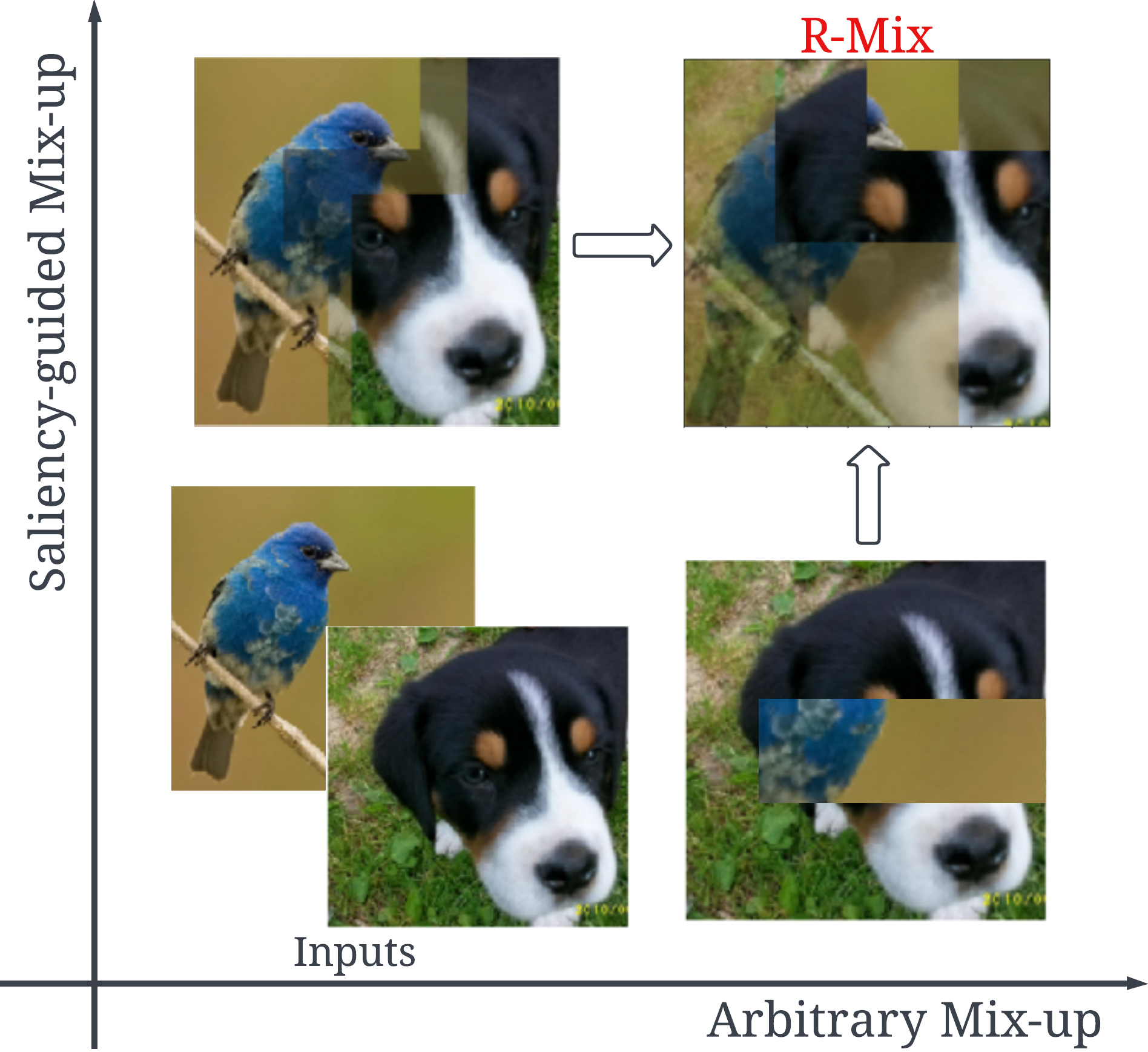}
  \caption{Illustration of our proposed method \rrlmix{}. Arbitrary Mix-up linearly interpolates images or employs a cut-and-paste strategy. Saliency-guided Mix-up preserves the rich supervisory signals of the images. Our method \rrlmix{} works by combining the finest aspects of both approaches and demonstrates its effectiveness on a variety of tasks.}
  \label{fig:overview}
\end{figure}

Among the rich family of mix-up extensions, 
a popular branch in it is mix-up methods 
that leverage the information of \emph{saliency maps}, 
because intuitively, 
one way to improve the efficiency of mix-up would be to replace its random procedure with a directed procedure guided by some additional knowledge, 
and a saliency map appears to be a natural choice for such knowledge. 

Probably driven by the same intuition, 
the community has investigated the saliency-based mix-up idea deeply 
in recent years,
such as \saliencymix{} \cite{uddin2021saliencymix}, \puzzlemix{} \cite{kim2020puzzlemix}, and \comixup{} \cite{kim2021comixup}. \saliencymix{}  generates the saliency map, and then employs the cut-and-paste strategy of \cutmix{}. \puzzlemix{} further introduces secondary optimization objectives that first optimize the saliency map, then optimize the transport plan in order to preserve the rich supervisory signals of the image. \comixup{} extends \puzzlemix{}'s idea by further introducing objectives to find the most suitable image to mix in the whole batch. 

We observe that each "direction" of image combining has its own advantages and disadvantages. \textbf{Arbitrary Mix-up} techniques, such as Input Mix-up and CutMix, offer fast training speed and simplicity while maintaining competitive performance. Contrarily, \textbf{Saliency-guided Mix-up}, like PuzzleMix and Co-Mixup, compromises speed and simplicity in favor of accuracy, expected calibration error, and robustness to adversarial attack. Over time, significant efforts have been proposed to further improve the saliency-guided direction with minimal focus on the other \cite{uddin2021saliencymix, kim2020puzzlemix, kim2021comixup, mixup9}, resulting in an unexplored randomness domain. We raise the question: \emph{is it feasible to have a method that is expeditious, simple, and effective at the same time?}

 In this paper, we identify a straightforward learning heuristic that sits in the middle of two paths. Our throughout the examination of saliency-guided methodologies suggests that they typically fall under a three-step optimization: First, calculate the saliency of the image. Then, mix the images in accordance with a secondary optimization objective. Finally, train the DNN with mixed images and labels. Roughly speaking, all three levels require the same amount of training time, making the training takes at least three times longer. We notice that, by swapping out the second step with a randomness-driven mixing approach, we are able to design a strategy that gives a competitive performance with state-of-the-art methods while maintaining the speed and ease of implementation of an arbitrary mix-up.

We name our method \textbf{\rrlmix{}} and empirically validate its performance on \textbf{four different tasks}: image classification, weakly supervised object localization, expected calibration error, and robustness to adversarial attack. On all benchmarks, \rrlmix{} shows an improvement or on-par performance with state-of-the-art methods. 

In summary, our contributions in this paper are as follows:
\begin{itemize}
    \item We begin by demonstrating that our implementation of arbitrary mix-up is capable of outperforming saliency-guided mix-up, indicating that existing attempts have not yet fully investigated the effectiveness of randomization. (Section: Background and Motivation).
    
    \item Motivated by the superiority of each mix-up direction over one another, we propose a novel method \rrlmix{} that combines the two mix-up routes and eliminates a third of the computational complexity (Section: R-Mix). With regard to four benchmarks on different model architectures: image classification, weakly supervised object localization, robustness to adversarial attack, and expected calibration error (Section: Experiments), we highlight that \rrlmix{} performs better or equally well as state-of-the-art approaches.
    
    \item Finally, to answer the question of whether coupled randomness and saliency are sufficient to the gain of \rrlmix{} or whether there exists a superior decision protocol, we present several experiments in the case of Reinforcement Learning controlled mix-up scenario. Our Reinforcement Learning agent adapts and chooses the mix-up rules based on the performance of the classifiers, aiming to reduce reliance on human-designed objectives and hyperparameter tuning of mix-up in general. We validate its effectiveness on \cifar{} image classification task and find that it performs competitively with other baselines. (Section: Ablation Studies).
\end{itemize}

%% file: secs/background.tex
\subsection{Mix-up Background}
Let $C, W, H, N$ denote the number of channels, image width, image height, and number of classes, respectively. We assume that $W = H$ for simplicity, and will use only $W$ from now on. Let $x \in \mathcal{X}, x \in \mathbb{R}^{C \times W \times W}$ be the input image and $y \in \mathcal{Y}, y \in \{0, 1\}^N$ be the output label. Let $f(\cdot;\theta_c)$ denote a classifier specified by parameter $\theta_c$. Let $\mathcal{D}$ be the distribution over $\mathcal{X} \times \mathcal{Y}$. In mix-up based data augmentation, the goal is to optimize the model's loss $\ell: \mathcal{X} \times \mathcal{Y} \times \Theta \rightarrow \mathbb{R}$ given the mix-up function for the inputs $h(\cdot)$, for the labels $g(\cdot)$, and the mixing distribution, usually $Beta(\alpha, \alpha)$ with the scalar parameter $\alpha$, as follows:

\begin{equation}
\label{objective}
    \underset{\theta}{\text{minimize}} \underset{(x_0, y_0), (x_1, y_1) \in \mathcal{D}}{\mathbb{E}} \underset{\lambda \sim Beta}{\mathbb{E}}\ell(h(x_0, x_1), g(y_0, y_1);\theta_c))
\end{equation}

Mix-up typically requires two tuples of images-labels. \mixup{} \cite{zhang2018mixup} defines $h(x_0, x_1) = \lambda x_0 + (1-\lambda) x_1$ and $g(x_0, x_1) = \lambda y_0 + (1-\lambda) y_1$. \manifoldmixup{} \cite{verma2019manifold} extends \mixup{} by mixing the inputs at a hidden representation $F$ as: $h(x_0, x_1) = \lambda F(x_0) + (1-\lambda) F(x_1)$, that is, at a random layer of $f(\cdot;\theta_c)$. \cutmix{} randomly copies a rectangular region from $x_0$ and pastes it to $x_1$. \puzzlemix{} ~\cite{kim2020puzzlemix} uses $h(x_0, x_1) = z \odot \Pi^T x_0 + (1-z) \odot \Pi'^T x_1$ where $\Pi$ is a transport plan, $z$ is a binary mask and $\odot$ is element-wise multiplication. \comixup{} \cite{kim2021comixup} extends $h(\cdot)$ to operate on a batch of data instead of two pairs: $h(x_B)$. 

While some early techniques simply mix the images using weights sampled from the $Beta$ distribution (\mixup{}, \manifoldmixup{}), or choose an arbitrary rectangular region and then apply mix-up (\cutmix{}), both \puzzlemix{} and \comixup{} require additional optimization objectives to ensure rich supervisory signals, introducing heavy computational cost.

\subsection{Motivation}
\label{sec:motivation}
In this Section, we provide empirical evidence demonstrating the usefulness of randomization on generalization ability. \cutmix{} \cite{yun2019cutmix}, which randomly cuts a rectangular patch from one image and pastes it into another, is typically justified on the grounds that it can create abnormal images by unintentionally choosing the fragments that do not contain any information about the source object (for example, cutting the grass-only region in an image of a cow on grass), which results in the so-called "learning false feature representations". Recent saliency-based methods aim to solve the issue by enhancing the saliency of the combined pictures, and report an increase in performance \cite{uddin2021saliencymix,kim2020puzzlemix,mixup9,kim2021comixup}. Naturally, one may think that saliency is the main contributing factor to this increment. However, we provide several counter-examples suggesting that this notion is only partially persuasive, as randomness is still essential for generalization.

\begin{table}[t!]
\centering
  \begin{tabular}{lcc}
    \hline
    Scheduler & MultiStepLR & OneCycleLR \\ \hline
    CutMix (300) & 78.71 & $\mathbf{80.60^\dagger}$\\ 
    PuzzleMix (300) & 79.38 & 79.75 \\ 
    Co-Mixup (300) & \textbf{80.13} & 79.79 \\ \hline
    PuzzleMix (1200) & 80.36 & 80.48 \\ 
    Co-Mixup (1200) &80.47 & 80.30 \\ \hline
  \end{tabular}
  
  \caption{Top-1 Accuracy on \cifar{} using PARN-18 with diffferent learning rate schedulers and training epochs. Bold indicates the best result. $\dagger$ denotes the \cutmix{}+ version we will use throughout this paper.}
  \label{tab:schedulersPARN}
\end{table}

\begin{itemize}
    \item \textbf{OneCycleLR elevates arbitrary-mixup to cutting-edge tier.} We simply change another LR scheduler instead of using the default MultiStepLR, specifically the OneCycleLR scheduler \cite{smith2018superconvergence} and reproduce \cutmix{}. We denote this method as \cutmix{}+. We train PreActResNet-18 (PARN-18) \cite{he2016preact} on \cifar{} for 300 epochs. From Table \ref{tab:schedulersPARN} (first row), \cutmix{}+ performs better than the most advanced saliency-guided mix-up by increasing accuracy by 1.89\%.
    Swapping another LR scheduler takes a few lines of code and introduces no additional computational cost.

    \item \textbf{OneCycleLR does not help saliency-guided mix-up.} We evaluate the performance of the OneCycleLR scheduler by using hyperparameters from the prior \cutmix{}+ experiment and reproduce these two saliency-guided methods again for 300 epochs. Table \ref{tab:schedulersPARN} (right column) demonstrates that OneCycleLR improves \puzzlemix{} by $0.37\%$ accuracy, but not \comixup{}.

    \item \textbf{Training saliency-guided mix-up for four times as many epochs still underperform \cutmix{}+}. We run \puzzlemix{} and \comixup{} for 1200 epochs using both schedulers and compare the results. Table \ref{tab:schedulersPARN} (last two rows) further shows that despite the improvement, both methods still fall short of \cutmix{}+. 
    
    \item \textbf{The finding is in line with other model architectures}. Finally, we further solidify the findings by running \cutmix{}+ on three more model architectures: WideResNet (WRN) 16-8 \cite{zagoruyko2017widern}, ResNeXt29-4-24 \cite{xie2016resnext} for 300 epochs, and WRN 28-10 for 400 epochs following the original implementations. Table \ref{tab:schedulers} shows that \cutmix{}+ consistently bests other state-of-the-art methods up to 2.08\% accuracy.
\end{itemize}

    \begin{table}[t]
\centering
  \begin{adjustbox}{width=1\columnwidth}
  \begin{tabular}{lccc|c}
    \hline
    Method & CutMix & PuzzleMix & Co-Mixup & CutMix+ \\ \hline
    PARN18 & 78.71 & 79.38 & 80.13 & \textbf{80.60} \\ 
    WRN16-8 & 79.86 & 80.76 & 80.85 & \textbf{81.79} \\ 
    WRN28-10 & 82.50 & \textbf{84.05} & - & 83.97 \\ 
    RNX & 78.14 & 78.88 & 80.22 & \textbf{82.30} \\ \hline
    \end{tabular}
    
    \end{adjustbox}
  
  \caption{Top-1 Accuracy (\%) on \cifar{} with various model architectures. Higher is better. Bold indicates the best result.}
  \label{tab:schedulers}
\end{table}

These empirical results hint that although \cutmix{}, and arbitrary mix-up in general, may produce distorted visuals, they are not as problematic in practice as they would appear. We then build a method with the following objectives as our driving force:
\begin{enumerate}
    \item \textbf{Experimenting with OneCycleLR scheduler}: \cutmix{} simply requires a different LR scheduler to perform better. While MultiStepLR clearly helps \puzzlemix{} and \comixup{}, OneCycleLR helps \cutmix{} and provides little or no performance benefit for the other two. In this work, OneCycleLR serves as the primary engine for our experiments.
    
    \item \textbf{Un-natural images help in generalization}: Complex mix-up techniques only produce marginal performance benefits when training for extended period of time, and \cutmix{} still remains a competitive method. Despite the fact that maximizing the saliency generates better-looking images for human eyes \cite{kim2020puzzlemix,kim2021comixup}, \cutmix{}+'s performance suggests that it may not be the optimal way to mix images. Instead, striking a balance between the most and least salient regions by combining randomness and saliency may yield a more promising outcome.
    
    \item \textbf{Low computational overhead}: Recent saliency-based mix-up algorithms have an excessively high computational cost. For instance, if all factors are held constant, \comixup{} takes 15 hours and \puzzlemix{} takes 27 hours, meanwhile Vanilla, \cutmix{} (and \cutmix{}+ variation) training takes approximately about 2.5 hours. Having an alternate mix-up method that compromises between simplicity, performance, and computing cost will tremendously benefit low-resource academic labs, businesses, and competitors in the data science field, where limited hardware is provided.
\end{enumerate}

%% file: secs/method.tex
In our proposed \rrlmix{}, we extend \mixup{} and \cutmix{} to the patch level but also utilize saliency information. We break down our method in four main steps.

\subsubsection{(1) Generating the Saliency Map:} First, we compute the saliency map $\phi(x)$ as the gradient values of training loss with respect to the input data and measure the $\ell_2$ norm across the input channels \cite{simonyan2014sal}:
\begin{equation}
\label{saliency}
    \phi(x) = \sqrt{\sum_{i=0}^C \frac{1}{C} (\nabla^i_{\theta_c}\ell (x, y;\theta_c))^2}
\end{equation}
where $\nabla^i_{\theta_c} \ell (x, y;\theta_c)$ denotes the gradient at channel $i$.

Second, we normalize $\phi(x)$ so that all elements of the map sum up to 1, and down-sample it to size $p \times p$, where $p$ is arbitrarily chosen as a multiple of 2. Intuitively, normalizing and down-sampling ensure numerical stability and decrease compute costs for the subsequent operations. Moreover, choosing a random $p$ for each batch enhances sample diversity. Specifically:
\begin{equation}
\label{downsample}
    \phi'(x) = \text{AvgPool}\left(\frac{\phi(x)}{\sum \phi(x)}, \text{kernel\_size} = p, \text{stride}=p \right)
\end{equation}

\subsubsection{(2) Splitting the Saliency Map into two regions:} Next, we randomly partition $\phi'(x)$ into two regions: the most and least salient regions, using the percentile value. For the top-k space $\mathcal{A}$ with $K$ equally spaced values from 0.0 to 0.99, we sample a value $q \in \mathcal{A}, q \in [0, 0.99]$. We compute the $q$-th percentile value of the down-sampled saliency map $\phi'(x)$, denoted as $q_{\text{perc}}$ and construct a binary mask $m$ as follows. For the $i$-th element in $\phi'(x)$:
\begin{equation}
\label{mask}
m(i)= 
    \begin{cases}
    1, & \text{if } \phi'_i(x) \geq q_{\text{perc}} ~(\textbf{top} \text{ salient region}) \\
    0, & \text{otherwise} ~(\textbf{least}  \text{ salient region})
    \end{cases}
\end{equation}

\begin{figure}[t]
  \centering
  \includegraphics[width=0.99\columnwidth]{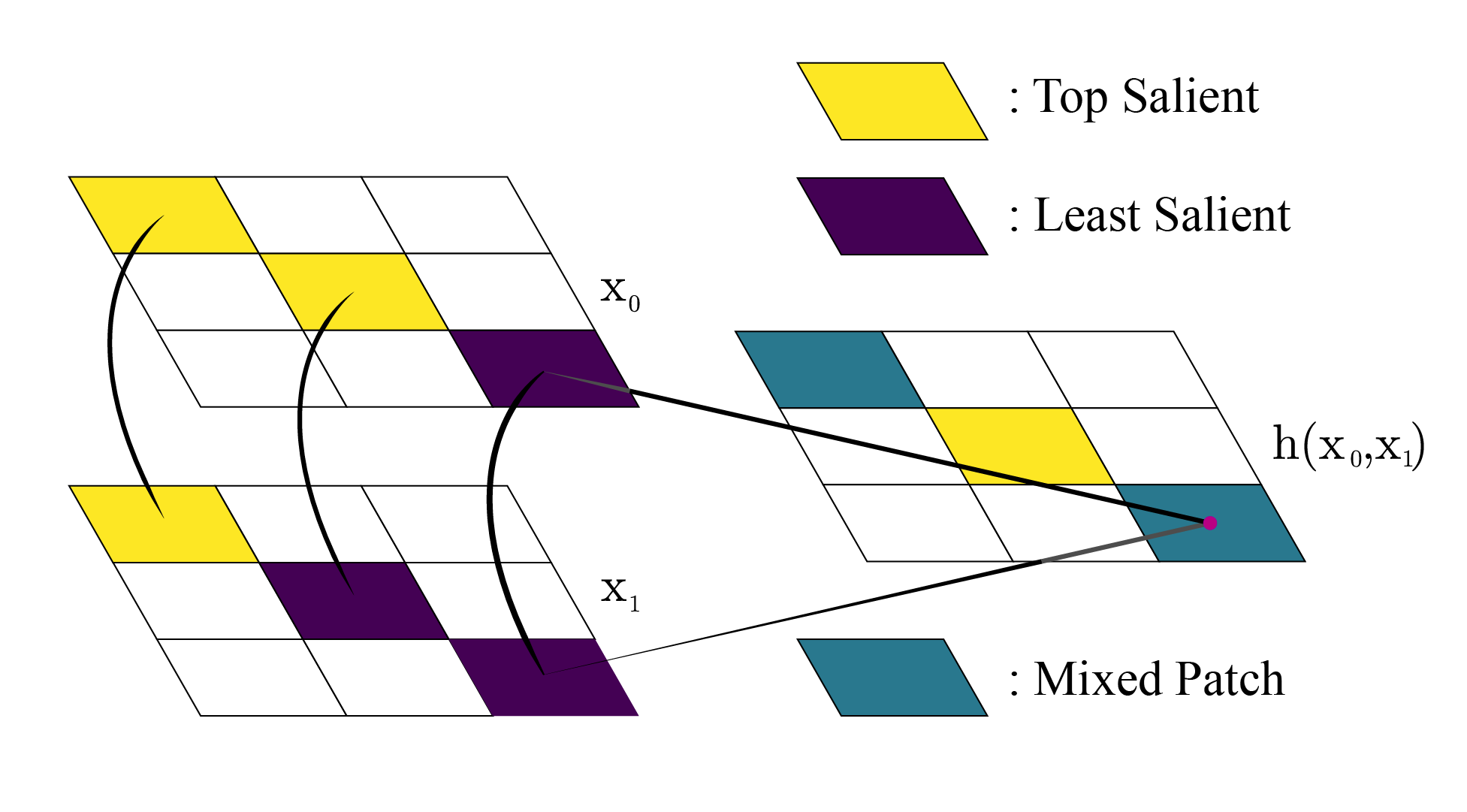}
  \caption{Illustration of R-Mix mix-up process. Given two images $(x_0, x_1)$ and their saliency maps, for two patches at the same position, if they both belong to the top (yellow) and least (purple) salient regions, we mix the patch (blue), else we only select the top salient one. Best viewed in color}
  \label{fig:mixup}
\end{figure}

After construction, mask $m$ is up-sampled by replicating elements to match the size of the inputs $x$. After this step, we obtain the mask of the top-least salient regions of the inputs.

\subsubsection{(3) Creating the soft mix-up filter:} For a pair of $(x_0, x_1)$, we obtain $(m_0, x_0)$ and $(m_1, x_1)$. We construct another mask $m_{\text{inter}}$ (inter stands for \textit{intersection}) as follows. For the $i$-th element in $m$:
\begin{equation}
\label{intersec}
    m_{\text{inter}} (i)= 
\begin{cases}
    1, & \text{if } m_0 (i) = m_1 (i)\\
    0,              & \text{otherwise}
\end{cases}
\end{equation}

The value of the element is $1$ if the two corresponding patches both belong to the top and least salient regions, and $0$ otherwise (Figure \ref{fig:mixup}).

\subsubsection{(4) Mixing the images and labels:} Finally, we sample the mixing coefficient $\lambda \sim Beta(\alpha, \alpha)$, then our mix-up function is defined as:
\begin{align}
\label{mixup}
\begin{split}
    h(x_0, x_1) & = m_{\text{inter}} \odot (\lambda x_0 + (1-\lambda) x_1) \\ 
    & + \neg m_{\text{inter}} \odot (m_0 \odot x_0 + m_1 \odot x_1)  
\end{split}
\end{align}
where $\neg$ denotes the logical NOT operator, that is, the binary mask is flipped. In short, for the $i$-th element in $m_0$ and $m_1$, we mix the element if $m_0 (i) = m_1(i)$ (analogous to \mixup{}). For the elements that $m_0 (i) \neq m_1(i)$, we use $m_{\text{inter}}(i) = \max(m_0(i), m_1(i))$ (analogous to \cutmix{}). Note that $m$ is a binary mask.

Let $c(m)$ denote the number of elements that are active, that is, $c(m) = |\{i|m(i) = 1\}|$ where $|\cdot|$ denotes the cardinality of a set. The label mix-up function is defined as:
\begin{align}
\label{mixup-label}
\begin{split}
    g(y_0, y_1) & =  \dfrac{c(m_{\text{inter}})}{W \times W}(\lambda y_0 + (1-\lambda) y_1) \\
    & + \dfrac{c(\neg m_{\text{inter}} \odot m_0)y_0 + 
    c(\neg m_{\text{inter}} \odot m_1)y_1}{W \times W}
\end{split}
\end{align}

This label mix-up function takes into account both the mix-up $\lambda$, and how many patches of the image are mixed.

In practice, all operations can operate on a batch level, with the current batch being randomly permuted to obtain the other input. The mixed sample produced by Equation \ref{mixup} and \ref{mixup-label} is used to train the classifier $f(\cdot;\theta_c)$ to minimize the soft target labels by minimizing the multi-label binary cross-entropy loss in Equation \ref{objective}.

We list the major distinctions between \rrlmix{} and alternative techniques in Table \ref{tab:differences}. In summary, our method \rrlmix{} has fast training speed, and utilizes the entire image, but has no additional optimization objective. Figure \ref{fig:mixup} further illustrates the mix-up process of two images on the patch level.

\begin{table}[t!]
\centering
\begin{adjustbox}{width=1\columnwidth}
\begin{tabular}{lcccc|c}
\hline
 & {Mix-up} & Cutout & CutMix & Co-Mixup & R-Mix \\ \hline
Speed & 1/1 & 1/1 & 1/1 & 3/1 & 2/1 \\
Top Salient & ? & ? & Mixed & Yes & Yes \\ 
Least Salient & ? & ? & Mixed & No & Yes \\ 
Full Image & Yes & No & Yes & Yes & Yes \\ 
Dropout & No & Yes & Yes & Yes & Yes \\
Mixed $(x,y)$ & Yes & No & Yes & Yes & Yes \\ 
$2^{\text{nd}}$ objective & No & No & No & Yes & No \\
\hline

\end{tabular}
\end{adjustbox}

\caption{Major distinctions between \rrlmix{} and other techniques. Training speed is measured on the same GPU.}
\label{tab:differences}
\end{table}

%% file: secs/experiment.tex
\begin{table*}[t]
  \centering
  \begin{tabular}{lcccc|cc|cc}
\hline
Method & Vanilla & Input & Manifold & CutMix & PuzzleMix & Co-Mixup & \textbf{CutMix+} & \textbf{R-Mix} \\ \hline
PARN-18 & 76.41 & 77.57 & 78.36 & 78.71 & 79.38 & 80.13 & 80.60 & $\mathbf{81.49}$ \\ 
WRN16-8 & 78.30 & 79.92 & 79.45 & 79.86 & 80.76 & 80.85 & 81.79 & $\mathbf{82.32}$ \\ 
WRN28-10 & 78.86 & 81.73 & 82.60 & 82.50 & 84.05 & - & 83.97 & $\mathbf{85.00}$ \\ 
RNX & 78.21 & 78.30 & 77.72 & 78.14 & 78.88 & 80.22 & 82.30 & $\mathbf{83.02}$ \\ \hline
\end{tabular}
\caption{Top-1 Accuracy (\%) on \cifar{} with various models and methods trained for 300 epochs. Higher is better. Bold indicates the best result.}
  \label{tab:results}
\end{table*}

\subsubsection{Datasets.} We test our methods on two standard classification dataset benchmarks. \cifar{} \cite{Krizhevsky09cifar100} contains 50k images of size $32 \times 32$ for training and 10k images for validation, equally distributed among 100 classes.

\imagenet{} \cite{Russakovski2015ImageNet} has 1.3M images for training distributed among 1k classes and has 100k images for validation. We normalize the data channel-wise, and average the results over 10 runs on \cifar{}, 5 runs on \imagenet{}. Similar to earlier works, traditional augmentations, such as Random Horizontal Flip and Random Crop with Padding, are employed. 

\subsubsection{Model Architecture.} To remain consistent with earlier works, we use five different model architectures to test our method. We use PreActResNet-18 (PARN18) \cite{he2016preact}, Wide Res-Net (WRN) 16-8 and 28-10 \cite{zagoruyko2017widern}, and ResNeXt 29-4-24 (RNX) \cite{xie2016resnext} on \cifar{}. For \imagenet{} we use ResNet-50 \cite{he2016resnet}.

\subsubsection{Pipeline and Hyperparameters.}
For \cifar{}, we set $p \in \{2, 4\}, K=10, \alpha=1.0$ and use OneCycleLR scheduler with initial LR $3e-3$, max LR $0.3$ and final LR $3e-5$, increasing for $30\%$ of the total number of epochs. We train for a total of 300 epochs with a batch size of 100. For \imagenet{} we use the identical protocol (such as image size and LR scheduler) described in \puzzlemix{} and \comixup{}, which trains ResNet-50 for 100 epochs. We set $p \in \{2, 4\}, K=10, \alpha=0.2$.

\subsection{Image Classification}
For fair comparison, we include results that were reported using the same training pipeline, that are: \mixup{} \cite{zhang2018mixup}, \manifoldmixup{} \cite{verma2019manifold}, \cutmix{} \cite{yun2019cutmix}, \puzzlemix{} \cite{kim2020puzzlemix}, \comixup{} \cite{kim2021comixup}, but add other methods with different pipelines for comparison in the Appendix. All methods are trained using PARN-18, WRN16-8, and RNX on \cifar{} for 300 epochs, except WRN28-10 is trained for 400 epochs.

\begin{table}[t!]
  \centering
  \begin{tabular}{lccc}
  \hline
    Metric & Accuracy  & Localization & Speed\\ \hline
    Vanilla & 75.97  & 54.36 & 1/1\\ 
    Input & 77.03  & 55.07 & 1/1\\
    Manifold & 76.70  & 54.86 & 1/1 \\ 
    CutMix & 77.08  & 54.91 & 1/1 \\ 
    PuzzleMix & 77.51 & 55.22 & 2.8/1 \\ 
    Co-Mixup & \textbf{77.61}  & 55.32 & 3/1 \\ \hline
    $\text{\rrlmix{}}$ & 77.39 & \textbf{55.58} & 2/1 \\ \hline
  \end{tabular}
  \caption{Top-1 Accuracy, Localization Accuracy (\%), and Training speed increment on \imagenet{} using ResNet-50 trained for 100 epochs. Higher is better. Bold indicates the best result.}
  \label{tab:resultsImgNet}
\end{table}

From Table \ref{tab:results}, \rrlmix{} outperforms \cutmix{} by $2\%$ and \cutmix{}+ by 1\% on average. It outperforms \comixup{} by $1.47\%$ with WRN16-8, by $0.85\%$ with WRN28-10 and by $2.8\%$ with RNX. As noted in other works \cite{zhang2018mixup}, mix-up methods generally benefit more from models with higher capacity, explaining the higher gain on bigger models.

We further test \rrlmix{} on \imagenet{} (ILSVRC 2012) dataset \cite{Russakovski2015ImageNet}. We use the same training protocol as specified in \comixup{} which trains ResNet-50 for 100 epochs. Table \ref{tab:resultsImgNet} shows that \rrlmix{} shows an improvement over Vanilla by $1.42\%$ and \cutmix{} by $0.31\%$.

\subsection{Weakly Supervised Object Localization}
Weakly Supervised Object Localization (WSOL) aims to localize an object of interest using only class labels without bounding boxes at training time. WSOL operates by extracting visually discriminative cues to guide the classifier to focus on prominent areas of the image.

We compare the WSOL performance of classifiers trained on \imagenet{} to demonstrate that, despite the fact that \rrlmix{} produces un-natural images, it is \emph{more effective} in focusing on salient regions compared to other saliency-guided methods. From Table \ref{tab:resultsImgNet}, using the Class Activation Map method \cite{zhou2015learning} and the protocol described in \comixup{}, it is interesting that, even with a lower Top-1 Accuracy, our method \emph{increases} the Localization Accuracy by $0.26\%$ and outperforms all other baselines. This further suggests that by striking a balance between the most and least salient regions, \rrlmix{} better guides the classifier to focus on salient regions.

\begin{table}[h!]
\centering
\begin{tabular}{lcc}
\hline
Metric & ECE & FGSM \\ \hline
Vanilla & 10.25 & 87.12 \\
Input & 18.50 & 81.30 \\
Manifold & 7.60 & 80.29 \\
CutMix & 18.41 & 86.96 \\
PuzzleMix & 8.22 & 78.70 \\ 
Co-Mixup & 5.83 & 77.61 \\\hline
\textbf{R-Mix} & \textbf{3.73} & \textbf{77.08} \\ \hline
\end{tabular}
\caption{Expected Calibration Error (ECE) (\%) and Top-1 Error Rate (\%) of PARN-18 to FGSM attack. Lower is better.}
\label{tab:ecefgsm}
\end{table}

\subsection{Expected Calibration Error}
We evaluate the expected calibration error (ECE) \cite{guo2017calibration} of PARN-18 trained
on CIFAR-100. ECE is calculated by the weighted average of the absolute difference between the confidence and accuracy of a classifier. From Table \ref{tab:ecefgsm}, we show that while Arbitrary Mix-up methods tend to have \emph{under-confident} predictions, resulting in higher ECE value, Saliency-guided Mix-up methods tend to have best-calibrated predictions. Our method \rrlmix{} successfully alleviates the over-confidence issue and does not suffer from under-confidence predictions.

\subsection{Robustness to Adversarial Attack}
Adversarial Attack attempts to trick DNNs into classifying an object incorrectly by applying small perturbations to the input images, resulting in an indistinguishable image for the human eye. \cite{szegedy2013intriguing}. Following previous evaluation protocol \cite{kim2020puzzlemix}, we evaluate PARN-18 model's robustness to FGSM adversarial attack with $8/255$ $\ell_{\infty}$ $\epsilon$-ball. As shown in Table \ref{tab:ecefgsm}, we observe that Saliency-guided methods have lower FGSM error. By leveraging this Saliency information, \rrlmix{} further establishes the best result among other competitors by lowering the Error Rate by 0.53\%. 

\subsection{Computational Analysis}
We compare the wall time on \cifar{} and \imagenet{} by investigating the released checkpoints and reproducing experiments. Specifically, including training and validation at each epoch, for \textbf{\cifar{}} with batch size 100, \comixup{} takes 15 hours on one RTX 2080Ti, whereas \rrlmix{} takes \textbf{4.0} hours. For \textbf{\imagenet{}} with 4 RTX 2080Ti, vanilla training takes 0.4s per batch, \rrlmix{} takes 0.77s per batch while \comixup{} takes 1.32s per batch. It should be noted that the saliency map is built on the gradient information \cite{simonyan2014sal} which requires two passes to the classifier. As a result, the running time is expected to be twice as long as with vanilla training. During validation, all classifiers need the same amount of time.

\section{Ablation Studies}
\label{sec:abl-study}
We conduct ablation studies about hyperparameter sensitivity and experiments about a mix-up method that automatically decides the mix-up policies based on the model's performance. We aim to lay the groundwork for future mix-up methods that require minimal human-designed objectives and low hyperparameter tuning effort.

\subsection{Sensitivity to Hyperparameters.}
\textbf{Number of patches $p$ and top-k space $K$.} We conduct hyperparameter tuning with different choices of the down-sampling Kernel Size $p$ and the top-k space that consists of $K$ equally-spaced values from $0.0$ to $0.99$ on \cifar{}. We then use the best found combination: $K=10, p \in \{2, 4\}$ to report the final result as in previous Tables and Figures. We report the result in Table \ref{tab:hyperparam}. We observe that, the higher the value $p$, the less efficient the method is. We hypothesize that, since each image patch has its own mix-up rule depending on the "other" patch, thus the higher the $p$ value, the higher the probability that a patch has different mixing rules compared to its neighbor patches. This diversity "breaks" the connectivity of the patches, which in turn hurts the convolution operations.

\begin{table}[h!]
\centering
\begin{tabular}{l|cccc}
 & K=5 & K=10 & K=16 & K=20 \\ \hline
$p \in \{2, 4\}$ & 81.22 & 81.49 & 81.35 & 81.18 \\
$p \in \{4, 8\}$ & 80.98 & 80.80 & 80.30 & 80.54 \\
$p \in \{8, 16\}$ & 79.94 & 80.33 & 79.29 & 79.50 \\
$p \in \{16, 32\}$ & 79.60 & 79.06 & 78.86 & 79.20 \\
$p \in \{2, 32\}$ & 79.64 & 79.69 & 79.21 & 79.64 \\
$p \in \{2, 4, 8\}$ & 80.87 & 80.34 & 80.04 & 80.94 \\
$p \in \{4, 8, 16\}$ & 80.12 & 80.37 & 80.34 & 80.41 \\
$p \in \{8, 16, 32\}$ & 79.53 & 79.12 & 79.55 & 79.67
\end{tabular}
\caption{Top-1 Accuracy on \cifar{} using PARN18 with different choices of hyperparameters. Higher is better.}
\label{tab:hyperparam}
\end{table}

\textbf{Mixing parameter $\alpha$.} We then conduct sensitivity analysis on the mixing parameter $\alpha$ used in sampling weights from the Beta distribution on CIFAR-100 with PARN-18 model. Table \ref{table:alpha} shows that for the majority of options, \rrlmix{} is still ourperforming other baselines and only suffers from minor accuracy lost, demonstrating its robustness to hyperparameter tuning.

\begin{table}[h!]
\centering
\begin{tabular}{lcccc}
\hline
 & $\alpha=0.2$ & $\alpha=0.5$ & $\alpha=1.0$ & $\alpha=2.0$ \\ \hline
R-Mix & 81.29 & 81.40 & 81.49 & 81.01 \\ \hline
\end{tabular}
\caption{Top-1 Accuracy of R-Mix on CIFAR-100 with different $\alpha$ values. Higher is better.}
\label{table:alpha}
\end{table}

\subsection{Is Randomness Enough? Reinforcement Learning-Powered Decisions with RL-Mix.}

In this section, we perform early experiments in an attempt to answer the question ``whether coupled randomness and saliency are sufficient to the gain of R-Mix or there exists a superior decision protocol`` using Reinforcement Learning. Inspired by AutoAugment \cite{Cubuk2019AutoAug}, we use the Proximal Policy Optimization \cite{schulman2017ppo} from Stable-Baselines3 \cite{stable-baselines3} using default hyperparameters suggested by a large-scale study \cite{andrychowicz2021whatmatter}. With the inputs as the saliency map $\phi'(x)$ and the logits, the agent determines the top-k value for each image in a batch. An episode of the agent ends when the classifier $f(\cdot, \theta_c)$ finishes training one epoch. Since the agent requires a fixed input size, we arbitrarily choose $p=8$. Based on the findings from \cite{zheng2022deep}, the reward function is the cosine similarity between the gradients of the original input $x$ and the mixed input $x'$, that is, $CosSim(\phi(x), \phi(x'))$. We call this method \textbf{RL-Mix}.

\begin{table}[h!]
  \centering
  \begin{tabular}{lcc|cc}
    \hline
    Model  & CutMix & CutMix+ & R-Mix & RL-Mix \\ \hline
    PARN-18  & 78.71 & 80.60 & \textbf{81.49} & 80.75 \\
    WRN16-8  & 79.86 & 81.79 & \textbf{82.32} & 82.16 \\
    WRN28-10  & 82.50 & 83.97 & \textbf{85.00} & 84.90 \\
    RNX & 78.14 & 82.30 & \textbf{83.02} & 82.43 \\ \hline
    \end{tabular}
  \caption{Top-1 Accuracy of RL-Mix on \cifar{} trained for 300 epochs. Higher is better.}
  \label{tab:results2}
\end{table}

Table \ref{tab:results2} reports the result of RL-Mix and other baselines. We can see that in most cases, \rrlmix{} is still better than RL-Mix. Interestingly, with a fixed size of $p$ and no hyperparameter tuning, it is still capable of delivering good performance. On the same GPU used throughout the paper, \textit{RL-Mix} is slower than \rrlmix{} by 2.0 times with a runtime of 7.5-8 hours.

Although RL-Mix is only early work, we believe it has the potential to open a new research direction of fully automatic mix-up, a branch in AutoML that requires minimal human-designed objectives and has low hyperparameter tuning effort.

%% file: secs/related.tex
\subsection{Saliency Maps}
There have been many works towards interpretability techniques for trained neural networks in recent years. Saliecny maps \cite{simonyan2014sal} and Class Activation Maps \cite{zhou2015learning} have focused on explanations where decisions about single images are inspected.
The work of \cite{simonyan2014sal} generates the saliency map directly from the DNN without any additional training of the network by using the gradient information with respect to the label. Following it, \cite{zhao2015sal} measures the saliency of the data using another neural network, and \cite{Zhou2016sal} aims to reduce the saliency map computational cost. We follow the method from \cite{simonyan2014sal}, which generates a saliency map without any modification to the model.

\subsection{Data Augmentation}
Data Augmentation is a technique to increase the amount of training data without additional data collection and annotation costs. There are two types of data augmentation techniques popularly used in various vision tasks: (1) transformation-based augmentation on a single image, and (2) mixture-based augmentation across different images. 

\subsubsection{Transformations on a single image.} Geometric-based augmentation and photometric-based augmentation have been widely used in computer vision tasks \cite{devries2017improved,huang2020self,huang2020improving,huang2020comprehensive, huang2022two, wang2022toward, wang2020high}. Survey papers \cite{Halevy2009data, Sun2017data, Shorten2019DataAugSurvey} show that inexpensive data augmentation techniques such as applying random flip, random crop, random rotation, etc., increase the diversity of the data and the robustness of the DNNs, and have been widely adopted in popular deep learning frameworks.

\subsubsection{Mixture across images.} 
\textbf{(i) Mixture of images with a pre-defined distribution.} \mixup{} \cite{zhang2018mixup} is a simple augmentation technique that blends two images by linearly interpolating them, and the labels are re-weighted by the blending coefficient sampled from a distribution. \manifoldmixup{} \cite{verma2019manifold} extends \mixup{} to the perturbations of embeddings. \cutmix{} \cite{yun2019cutmix} randomly copies a rectangular-shaped region of an image, and pastes it to a region of another image; \textbf{(ii) Mixture through Saliency Maps.} Saliency-based mixtures, such as \puzzlemix{} \cite{kim2020puzzlemix}, \comixup{} \cite{kim2021comixup}, and \saliencymix{} \cite{uddin2021saliencymix} first generate a saliency map, and then use the map to optimize secondary objective functions that maximize the saliency to mix the images and ensure reliable supervisory signals.

For a more comprehensive summary of recent mix-up methods \cite{mixup1, mixup2, mixup4, mixup6, mixup9, mixup10, mixup12, mixup13}, we refer readers to the survey paper \cite{naveed2021survey}.

\subsection{Deep Neural Networks Training Techniques}
Techniques such as Weight Decay \cite{Goodfellow-et-al-2016}, Dropout \cite{dropout}, Batch Normalization \cite{ioffe2015batch}, and Learning Rate schedulers are widely used to efficiently train deep networks.
The literature of learning rate (LR) scheduler is now nearly as extensive as that of optimizers \cite{pmlr-v139-schmidt21a}. Generally, the training is divided into multiple phases. The LR of the classifier is kept constant during a phase and then is decayed by a positive value in the next phase. One of the most common schedulers is MultiStepLR \cite{Goodfellow-et-al-2016,zhang2021dive} or step-wise decay, which divides the training into phases where each consists of tens or hundreds of epochs.
OneCycleLR, introduced in \cite{smith2018superconvergence} employs the cyclic learning rate scheduler \cite{smith2017cyclical} but only for one cycle. The LR starts with a small value, increases to the max value then gradually decreases to an even smaller value until training finishes.

In this paper, we show that the LR scheduler can have a large impact on the performance of existing mix-up methods, sometimes removing any performance gains of more sophisticated mix-up strategies compared to vanilla mix-up strategies.